%% file: main.tex
\documentclass[letterpaper]{article} 
\usepackage{aaai2027}  
\usepackage[hyphens]{url}  
\usepackage{graphicx} 
\usepackage{natbib}  
\usepackage{caption} 
\usepackage{algorithm}
\usepackage{algorithmic}

\usepackage{newfloat}
\usepackage{listings}
\DeclareCaptionStyle{ruled}{labelfont=normalfont,labelsep=colon,strut=off} 
\floatstyle{ruled}
\newfloat{listing}{tb}{lst}{}
\floatname{listing}{Listing}

\usepackage{booktabs}

\usepackage{multirow}
\usepackage{amssymb}
\usepackage{amsmath}

\title{AeroDPO: Unleashing Lightweight UAV Navigation with High-Fidelity Perception and Automated Preference Optimization}

\author {
    Peng Xu\textsuperscript{\rm 1,\rm 2}\equalcontrib,
    Chengcheng Wang\textsuperscript{\rm 1,\rm 2}\equalcontrib,
    Shaohua Wan\textsuperscript{\rm 1,\rm 2}\corresponding
}
\affiliations {
    \textsuperscript{\rm 1}University of Electronic Science and Technology of China, Chengdu, China\\
    \textsuperscript{\rm 2}Shenzhen Institute for Advanced Study, UESTC, Shenzhen, China\\
    \{xupeng23, wcc\}@std.uestc.edu.cn, shaohua.wan@uestc.edu.cn
}

\nocopyright

\begin{document}

\maketitle

\begin{abstract}

Vision-Language Navigation for Unmanned Aerial Vehicles (UAV-VLN) requires rapid and reactive control in complex 3D environments. Recent minimalist end-to-end paradigms show great promise but typically rely on massive language models containing billions of parameters, incurring prohibitive latency for real-world edge deployment. In this paper, we challenge this parameter-heavy reliance. Comprehensive cross-scale evaluations reveal the critical insight that perception quality fundamentally outweighs language reasoning capacity. We demonstrate that a lightweight 2B model equipped with high-fidelity visual inputs completely matches the overall success rates of massive 7B baselines. However, this minimalist policy exposes a fundamental robustness flaw inherent to pure Behavior Cloning (BC). Lacking explicit negative feedback, the agent fails to internalize robust spatial constraints and exhibits alarming collision rates in out-of-distribution (OOD) scenarios. To overcome this vulnerability without relying on unscalable human annotations, we propose AeroDPO, a zero-cost automated Direct Preference Optimization pipeline driven by deterministic physical simulation state rollback. Upon detecting collisions, the system autonomously rewinds the environment to extract causal reasoning errors as rejected actions, applies decoupled privileged interventions to synthesize collision-avoidance preferred maneuvers, and leverages an offline vision language inspector to filter visual ambiguities. By equipping our 2B model with this automated data flywheel, AeroDPO boosts success rates to 49.16\% on unmapped scenarios while drastically suppressing collision rates, establishing a new SOTA for autonomous aerial agents. Code is available at: \url{https://github.com/XuPeng23/AeroDPO}

\end{abstract}


\section{Introduction}

As a cornerstone of aerial embodied intelligence, the Unmanned Aerial Vehicle Vision-Language Navigation (UAV-VLN) task demands sophisticated capabilities ranging from precise environmental perception to real-time, reasoning-driven decision making \cite{liu2023Aerialvln,sapkota2025uavs, yao2025aeroverse}. To achieve this, mainstream methods heavily prioritize reasoning capacities, often resorting to complex hierarchical architectures or integrating external auxiliary modules, such as specialized memory mechanisms \cite{tang2026mitigating}. In contrast, recent pioneering works demonstrate that a simple, reactive policy can effectively replace these complex modular pipelines. For instance, AeroVLA \cite{xu2026aerovla} proposes a minimalist end-to-end Vision-Language-Action (VLA) framework capable of navigating open-world 3D spaces using merely fuzzy directional hints. However, despite such architectural simplifications, current state-of-the-art VLA paradigms inevitably rely on massive language model backbones \cite{wang2025uavflow}. This reliance incurs prohibitive inference latency and unacceptable power consumption that severely conflicts with the high-frequency reactive control demanded by dynamic aerial environments \cite{tian2025uavs, javaid2024large}. Consequently, breaking this parameter-heavy reliance is imperative for deploying highly agile aerial agents in the real world.

To overcome this deployment bottleneck, we first investigate the fundamental drivers underlying the success of minimalist VLA policies. Through comprehensive cross-scale evaluations, we reveal the critical insight that high-fidelity spatial perception fundamentally outweighs language reasoning capacity in UAV-VLN tasks. We demonstrate that an ultra-lightweight 2B model equipped with high-resolution visual inputs can effectively match the navigation success rate of massive 7B baselines. However, resolving the latency issue immediately exposes a crucial robustness vulnerability. Despite achieving superior task completion, the lightweight model trained via pure Behavior Cloning (BC) exhibits exceptionally high collision rates in out-of-distribution (OOD) environments \cite{wang2025uavflow}. Due to the inherent covariate shift and compounding errors in sequential decision-making \cite{ross2010efficient}, solely mimicking expert demonstrations without explicit negative feedback causes the agent to fail in internalizing robust structural boundaries, inevitably leading to crashes when encountering unfamiliar obstacles \cite{ross2011dagger}.

\begin{figure*}[t]
\centering
\includegraphics[width=\linewidth]{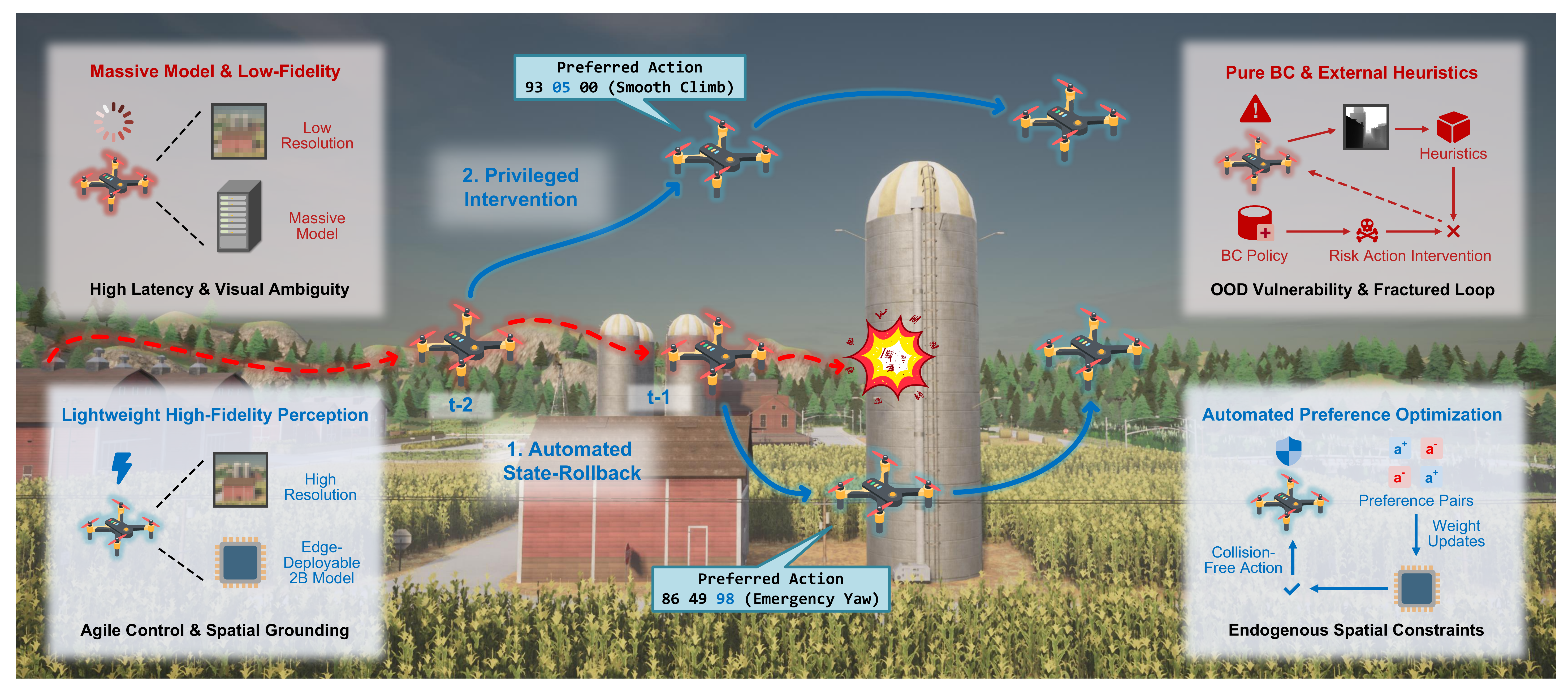}
\caption{Overview of the AeroDPO paradigm. Unlike parameter-heavy baselines and pure BC policies prone to collisions (red), AeroDPO leverages an ultra-lightweight 2B model. We internalize spatial awareness by exploiting automated state-rollback ($t-1$ and $t-2$) and privileged intervention to synthesize collision-free preferred actions (blue).}
\label{fig:figure_teaser}
\end{figure*}

While Direct Preference Optimization (DPO) \cite{rafailov2023direct} holds promise for spatial alignment, applying it to continuous aerial control faces the prohibitive cost of human annotations and the challenge of temporal credit assignment \cite{arjona2019rudder}. Since a crash is often the delayed consequence of an earlier maneuver, penalizing the immediate collision frame is inherently ineffective. To overcome these hurdles, we propose AeroDPO as a zero-cost automated offline preference alignment pipeline, as illustrated in Figure~\ref{fig:figure_teaser}. Instead of relying on human interventions or online heuristic patches, AeroDPO constructs an automated data flywheel comprising three core mechanisms. First, upon detecting a collision, the system leverages deterministic physical simulation state-rollback to isolate the upstream causal error as the rejected action. Second, a decoupled privileged intervention utilizes ground-truth spatial geometry to synthesize a collision-free preferred action. Finally, to guarantee optimization quality, an offline vision-language inspector rigorously filters these synthetic pairs to eliminate visually ambiguous scenarios. This automated data flywheel empowers the lightweight agent to intrinsically absorb structural boundaries without compromising flight fluidity.

Ultimately, by equipping the ultra-lightweight agent with our automated DPO pipeline, we rectify the inherent inefficiencies of pure BC and comprehensively outperform much larger baselines. The main contributions of this work are summarized as follows:

\begin{itemize}
    \item \textbf{Empirical Insight on Perception and Reasoning.} We demonstrate through comprehensive cross-scale evaluations that high-fidelity visual perception outweighs language parameter scale for minimalist UAV navigation.

    \item \textbf{Zero-Cost Automated DPO Pipeline.} We propose a novel preference mining framework that leverages simulation state rollback and privileged intervention to autonomously generate high-quality chosen and rejected maneuver pairs without human annotation.

    \item \textbf{Ultra-Lightweight SOTA Agent.} We deliver a highly efficient 2B-parameter UAV agent that internalizes robust spatial alignment and establishes new state-of-the-art performance on the TravelUAV \cite{wang2025towards} benchmark with minimal inference overhead.
\end{itemize}

\input{table_perception_reasoning}

\section{Related Works}

\subsection{UAV VLN and the Scalability Bottleneck}

The landscape of UAV-VLN is undergoing a rapid paradigm shift from hierarchical pipelines \cite{zhang2025citynavagent, chen2025typefly} to integrated VLA architectures \cite{chen2026vision, xia2026vision}. While this paradigm shows great promise, current methods struggle to balance model capacity with architectural simplicity, creating a critical scalability bottleneck. For instance, recent end-to-end models like AeroVLA \cite{xu2026aerovla} map raw visual inputs to control signals but still rely on a 7B model. The prohibitive inference latency of such massive models contradicts the high-frequency control demanded by aerial environments. Conversely, lightweight approaches like RaceVLA \cite{serpiva2025racevla} achieve deployment efficiency but remain confined to structured racing scenarios. To compensate for reasoning deficiencies or high latency, recent efforts regress to modular structures. CognitiveDrone \cite{lykov2025cognitivedrone} integrates an auxiliary Vision-Language Model (VLM) for cognitive capabilities, while LongFly \cite{jiang2025longfly}, See-and-Reach \cite{xue2026see}, and AeroDuo \cite{wu2025aeroduo} rely on external spatial trackers to achieve specific functions like precise landing. To bypass hardware limits, other works resort to compromise solutions, such as employing low-level operator fusions for marginal speedups \cite{wu2025vla}, relying on frozen models for training-free navigation \cite{hu2025see, liu2026indooruav}, or offloading computational burdens to remote servers \cite{zhao2026worldvln}. Consequently, returning to ultra-lightweight and fully integrated architectures represents the most viable path toward highly agile and edge-deployable aerial agents.

\subsection{The Robustness Dilemma of BC in Continuous Control}

Beyond computational bottlenecks, end-to-end aerial navigation faces a fundamental robustness dilemma. A majority of contemporary end-to-end models heavily depend on BC as their primary training paradigm. This pure imitation paradigm exhibits alarming brittleness in OOD scenarios, causing severe trajectory drift and catastrophic collisions \cite{xia2026vision, wang2025uavflow}. To mitigate this, conventional continuous control literature frequently patches policies with online collision-avoidance interventions. These range from Artificial Potential Fields (APF) \cite{khatib1986real, Alhaddad2024neural} and formal shielding frameworks \cite{alshiekh2018safe, dalal2018safe} to geometric heuristic correctors based on depth maps \cite{wu2025vla} and independent recovery networks \cite{thananjeyan2021recovery}. While preventing immediate collisions, these external patches severely compromise the unified perception-action loop of the end-to-end paradigm. Consequently, a truly robust embodied agent necessitates internalizing its spatial constraints, motivating a critical shift from online external patching to offline preference optimization.

\subsection{Preference Optimization in Embodied Agents}

Building upon foundational paradigms of learning from human feedback \cite{christiano2017deep}, Reinforcement Learning (RL) and Direct Preference Optimization \cite{rafailov2023direct} have shown immense potential in aligning complex robotic behaviors \cite{chen2026vision}. Nonetheless, applying preference optimization to continuous aerial control introduces severe challenges: the prohibitive cost of human annotations \cite{wang2025uavflow, wang2025deployable} and the intractable temporal credit assignment problem. While recent works incorporate RL fine-tuning into UAV navigation, their reward designs predominantly target static visual question answering or spatial bounding-box matching, failing to address the temporal alignment of continuous 3D flight trajectories. Breaking this impasse, and drawing inspiration from privileged learning paradigms \cite{chen2020learning, Mosbach2025prompt}, AeroDPO bypasses the credit assignment bottleneck via an annotation-free, simulation-rollback mechanism, autonomously synthesizing precise preference pairs to intrinsically absorb physical geometric boundaries.

\section{Empirical Insight into VLA Scaling for Aerial Navigation}

Before introducing our automated alignment pipeline, we conduct a comprehensive empirical analysis to dissect the true bottlenecks of end-to-end aerial navigation. Existing paradigms operate under the prevailing assumption that resolving complex aerial environments intrinsically necessitates scaling up the language reasoning backbone. This assumption forces a widespread reliance on massive models with billions of parameters or high-latency online application programming interfaces. To directly challenge this consensus, we construct a rigorous orthogonal evaluation on the TravelUAV benchmark \cite{wang2025towards}. By deploying the mainstream OpenVLA-7B \cite{kim2025openvla} and ultra-lightweight Qwen3-VL variants \cite{bai2025qwen3} on a single RTX 3090 GPU, we systematically isolate the impact of language model scale from perceptual fidelity. Ultimately, this analysis reveals that high-fidelity spatial perception fundamentally outweighs parameter-heavy reasoning while simultaneously exposing the severe robustness degradation inherent in pure BC.

\subsection{Deconstructing the Scaling Law between Perception and Parameter Scale}

\begin{figure}[t]
\centering
\includegraphics[width=0.95\linewidth]{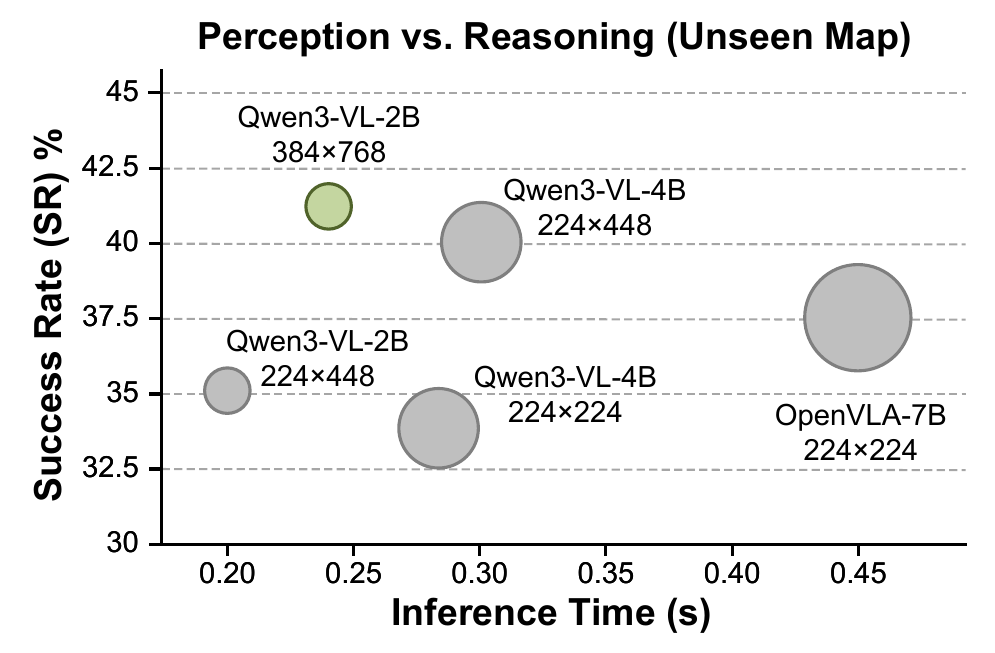}
\caption{Pareto frontier of perception versus parameters. The Qwen3-VL-2B model with $384{\times}768$ input establishes an optimal trade-off between Success Rate and Inference Time.}
\label{fig:figure_pareto}
\end{figure}

As detailed in Table~\ref{tab:perception_reasoning}, evaluating models under constrained square inputs ($224{\times}224$) demonstrates that the OpenVLA-7B outperforms the Qwen3-VL-4B. This dynamic validates that massive parameter scales provide robust reasoning capacity to compensate for visual ambiguity. However, processing an expanded resolution ($224{\times}448$) allows the 4B model to decisively surpass the 7B baseline. Furthermore, scaling the language backbone down to 2B parameters at this identical resolution predictably induces a performance drop. Strikingly, equipping this 2B model with a high-fidelity visual input of $384{\times}768$ directly recovers a performance level comparable to the 4B architecture and even comprehensively overtakes it on the highly challenging Unseen Map split. While this high-resolution 2B model may not strictly dominate across every map category, its exceptionally low computational and storage overhead grants it an absolute advantage for real-world edge deployment. As illustrated in Figure~\ref{fig:figure_pareto}, this configuration achieves a Pareto optimal solution for zero-shot generalization in unmapped environments. It secures the highest Success Rate (SR) while maintaining an ultra-low 5GB VRAM footprint and an agile 0.24s inference latency, perfectly balancing robust task completion with the high-frequency control demanded by unconstrained aerial navigation.

\subsection{The Robustness Vulnerability of Pure BC}

Although integrating high-fidelity visual inputs effectively neutralizes the task completion deficit, a granular analysis of the Collision Rate (CR) uncovers a critical vulnerability within the baseline policy. As detailed in Table~\ref{tab:perception_reasoning}, despite achieving a competitive 41.23\% Success Rate in unseen environments, the optimized 2B model exhibits an alarming CR of 46.76\%. This severe robustness degradation is an inherent artifact of the BC objective. By strictly minimizing behavioral divergence from expert demonstrations without explicit negative feedback, the pure imitation paradigm merely memorizes optimal paths under perfect conditions. The agent consequently fails to synthesize recovery strategies for inevitable trajectory drifts, rendering it exceptionally brittle against novel obstacles and devoid of robust internal spatial boundaries. This vulnerability dictates that perceptual enhancements alone cannot guarantee deployment readiness. Encoding intrinsic collision avoidance within the lightweight architecture, a fundamental paradigm shift from passive imitation to active preference optimization is imperative.

\section{The AeroDPO Framework}
\label{sec:method}

To eliminate the reliance on unscalable human annotations and instill endogenous spatial constraints into the end-to-end VLA paradigm, we propose AeroDPO. This framework transitions the ultra-lightweight agent from passive BC to active preference alignment via a zero-cost automated data flywheel.

\subsection{Kinematic-Preserving Rollback}
\label{subsec:rollback}

The automated data collection initiates during the rollout of evaluation trajectories in the simulator. When a collision is registered, penalizing the terminal impact state is kinematically meaningless due to aerodynamic inertia, as the agent has inevitably entered an irrecoverable physical deadzone, rendering any standard reactive recovery attempts completely futile and counterproductive. To capture the precursor actions responsible for accumulating collision risks, we implement a multi-scale \textit{Kinematic-Preserving Rollback} mechanism. The simulator rewinds to upstream waypoints $t-K$ ($K \in \{1, 2\}$), representing discrete spatial horizons for short-range emergency and long-range anticipatory avoidance. At these junctures, the flawed execution is designated as the rejected action $a^-$.

To explicitly safeguard the initial takeoff stabilization and final precision-landing phases from conflicting avoidance gradients, we enforce a strict trajectory-phase mask:
\begin{equation}
\mathcal{M}_{\text{filter}} = \mathbb{I}(t > 5) \cdot \mathbb{I}(\|p_{t-K} - p_{\text{target}}\|_2 \ge 15.0).
\end{equation}
where $\mathbb{I}(\cdot)$ denotes the indicator function, $t$ is the current episode step, $p$ represents the 3D spatial position, and $\|\cdot\|_2$ denotes the Euclidean distance. Only valid steps satisfying $\mathcal{M}_{\text{filter}} = 1$ are retained to extract the rejected action $a^-$ alongside the multimodal contextual state $o_{t-K}$. This state intrinsically couples the linguistic instruction and dual-view RGB images requisite for the downstream preference optimization with the aligned depth maps strictly reserved for the imminent APF computation.

\begin{figure*}[t]
\centering
\includegraphics[width=1\linewidth]{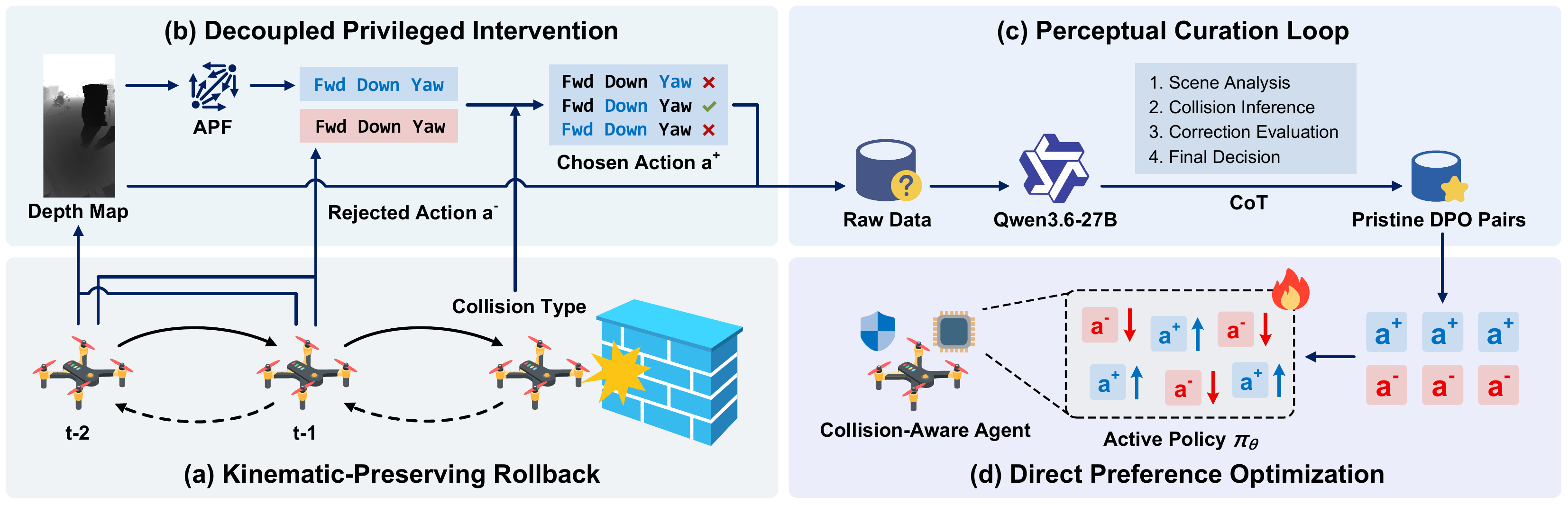}
\caption{The AeroDPO pipeline. (a) Kinematic-Preserving Rollback: Rewinding from collisions to extract rejected actions ($a^-$). (b) Decoupled Privileged Intervention: Leveraging depth-based APF to selectively correct specific axes and synthesize preferred actions ($a^+$). (c) Chain-of-Thought Perceptual Curation: Filtering visually ambiguous scenarios via offline VLM reasoning. (d) Direct Preference Optimization: Optimizing the model with curated pairs to internalize spatial constraints.}
\label{fig:figure_pipeline}
\end{figure*}

\subsection{Holonomic-Decoupled Privileged Intervention}
\label{subsec:intervention}

Having isolated the multimodal state $o_{t-K}$ and the rejected action $a^-$, we synthesize the optimal preferred action $a^+$ via a privileged APF. Leveraging exclusively the depth maps from $o_{t-K}$, we compute the attractive force toward the target and integrated repulsive forces from obstacles. The theoretical optimal action vector $a_{\text{apf}} = \langle \Delta x_{\text{apf}}, \Delta z_{\text{apf}}, \Delta \psi_{\text{apf}} \rangle$ is then projected. 

Unlike traditional APFs that indiscriminately override navigation intent, our \textit{Directional Dimension Filter} enforces granular interventions to guarantee evasion while maximally adhering to the original long-horizon routing semantics. The preferred action $a^+$ selectively integrates components of $a_{\text{apf}}$ based on the triggered sensor topology. Downward and lateral proximity warnings exclusively trigger corrections on the altitude ($\Delta z$) and heading ($\Delta \psi$) axes respectively. For frontal impacts requiring dynamic selection between horizontal circumvention and vertical traversal, we determine the dominant evasive axis $d^*$ by maximizing the normalized kinematic deviation:

\begin{equation}
d^* = \arg\max_{d \in \{z, \psi\}} \frac{| \Delta d_{\text{apf}} - \Delta d^- |}{\text{max}_d}.
\end{equation}
where $\text{max}_d$ represents the physical boundary of axis $d$. If altitude adjustment ($d^* = z$) is optimal for frontal evasion and the deceleration demand exceeds a threshold ($|\Delta x_{\text{apf}} - \Delta x^-| > 2$), the forward axis is concurrently overridden to prevent high-speed ascending collisions. Unselected axes strictly inherit their values from $a^-$.

\subsection{Chain-of-Thought Perceptual Curation}
\label{subsec:curation}

Synthetic data from privileged interventions inevitably introduces visual ambiguities (e.g., cluttered geometry or camera blinding), polluting the preference distribution. We rectify this via an offline VLM-supervised curation mechanism (Qwen3.6-27B). Given the multimodal state $o_{t-K}$ and synthesized action $a^+$, it executes a four-step autoregressive analysis: scene structural analysis, collision target identification, evasion verification, and final binary judgement. This eliminates suboptimal or physically unsolvable trajectories, ensuring only pristine preference pairs $(o_{t-K}, a^+, a^-)$ enter the optimization pool.

\subsection{Direct Preference Optimization}
\label{subsec:dpo}

Finally, we integrate the curated preference pairs into the DPO framework to internalize physical geometric boundaries directly into the policy weights. Following the tokenization scheme of AeroVLA, the optimization loss is computed exclusively over the 3-DoF control tokens by masking visual and textual prompts. For notational brevity, let $o$ denote the multimodal state $o_{t-K}$. We formulate the objective as follows:
\begin{equation}
\begin{split}
\mathcal{L}_{\mathrm{DPO}}(\pi_\theta; \pi_{\mathrm{ref}}) &= -\mathbb{E}_{(o, a^+, a^-)} \Bigg[ \log \sigma \Bigg( \beta \log \frac{\pi_\theta(a^+|o)}{\pi_{\mathrm{ref}}(a^+|o)} \\
&\quad - \beta \log \frac{\pi_\theta(a^-|o)}{\pi_{\mathrm{ref}}(a^-|o)} \Bigg) \Bigg]
\end{split}.
\end{equation}
where $\pi_\theta$ is the active VLA policy, $\pi_{\mathrm{ref}}$ represents the frozen baseline Supervised Fine-Tuning (SFT) policy, $\sigma$ denotes the logistic sigmoid function, and $\beta$ scales the divergence margin. This alignment bypasses complex reward engineering, enabling the agent to acquire error-correction capabilities in unfamiliar 3D environments.

\section{Experiments}
\label{sec:experiments}

\subsection{Experimental Setup}

\textbf{Benchmark and Datasets.} We evaluate our framework on the TravelUAV benchmark \cite{wang2025towards}, adhering strictly to the official Seen, Unseen Object, and Unseen Map test sets. The baseline SFT policy is trained on the AeroVLA expert dataset \cite{xu2026aerovla}. To construct the AeroDPO dataset, we deploy the baseline policy across 7,611 training routes, initially harvesting 3,855 raw preference pairs from 1,954 collision trajectories. The VLM curation loop subsequently filters these down to 2,005 pristine collision-avoidance pairs, yielding a final retention rate of 52.0\% for training.

\input{table_seen}

\input{table_unseen}

\textbf{Evaluation Metrics.} Following standard autonomous navigation protocols, we report Navigation Error (NE), Success Rate (SR), Oracle Success Rate (OSR), and Success weighted by Path Length (SPL). To explicitly quantify the efficacy of our preference optimization in establishing reliable collision-avoidance behaviors, we incorporate Collision Rate (CR) as a primary metric, defined as the percentage of episodes terminating prematurely due to critical physical intersections with the environment.

\textbf{Implementation Details.} We instantiate our agent using the Qwen3-VL-2B backbone \cite{bai2025qwen3}, scaling the visual input to $384{\times}768$. For automated preference curation, we deploy Qwen3.6-27B (4-bit quantized Q4\_K\_M) via \texttt{llama.cpp}. The VLM inference latency averages approximately 8 seconds per candidate pair, taking roughly 8.5 hours to process the entire corpus of 3855 raw trajectories. For the model architecture update, we apply LoRA \cite{hu2022lora} ($r=64$, $\alpha=128$, dropout=0.05) to the language backbone and fully fine-tune the visual projection layer. The DPO stage is trained over 2 epochs using AdamW with a learning rate of $2{\times}10^{-5}$, a KL penalty $\beta$ of 0.2, and a global batch size of 64 (micro-batch size of 4 with 16 accumulation steps). Crucially, the entire pipeline including AirSim \cite{Shah2018AirSim} data collection, VLM data curation, DPO training, and closed-loop model evaluation is executed seamlessly on a single NVIDIA RTX PRO 4500 Blackwell GPU.

\subsection{Evaluation on Seen Environments}

Table~\ref{tab:comparison_seen} summarizes the navigation performance on the Test Seen set. We first evaluate this split to verify whether the agent preserves fundamental flight capabilities after preference alignment. AeroDPO achieves a remarkable 60.93\% SR on the Full split, comprehensively outperforming the SFT Baseline (48.59\% SR) across all metrics. Specifically, it drives the NE down substantially from 69.14 to 53.30 while actively suppressing the CR from 34.27\% to 26.80\%. Notably, compared to prior state-of-the-art architectures (e.g., NavFoM, LongFly) that heavily rely on dense L1 Assist, AeroDPO demonstrates superior navigational efficiency (achieving 48.47\% SPL) using only coarse Fuzzy Direction. This confirms that our DPO pipeline successfully enables the model to learn complex autonomous obstacle avoidance while strictly maintaining its foundational cruising and landing capabilities. The alignment process significantly boosts overall task completion and collision-avoidance capabilities without triggering catastrophic forgetting.

\input{table_ablation}

\subsection{Zero-Shot Generalization on OOD Environments}

Table~\ref{tab:comparison_ood} details the zero-shot generalization results across OOD test sets. For the Unseen Object split, AeroDPO maintains highly robust performance, achieving a 67.41\% SR and an impressive 57.54\% SPL on the Full tasks. Notably, it secures a 7.63\% absolute SR improvement over the SFT baseline while driving the Navigation Error (NE) down to 47.31. This indicates that our endogenous collision-avoidance alignment seamlessly integrates with the innate open-vocabulary understanding of the VLM backbone, enabling the agent to efficiently evade novel object categories without targeted retraining.

The critical stress test, however, lies in the Unseen Map split, where pure BC policies typically suffer from severe compounding errors. Here, AeroDPO delivers a commanding 49.16\% SR on the Full split and an impressive 44.84\% SR on the long-horizon Hard tasks. In stark contrast, specialized models like LongFly plummet to an 11.27\% SR, exposing the fragility of heavily guided frameworks in unfamiliar 3D spaces. Furthermore, AeroDPO yields a 7.93\% absolute SR over the SFT Baseline while explicitly reducing NE. These results compellingly demonstrate that our automated DPO framework empowers the agent to internalize generalized structural boundaries rather than merely memorizing specific geometric features from the training maps.

\subsection{Ablation Studies}
\label{subsec_ablation}

Table~\ref{tab:ablation} comprehensively validates our core design choices, analyzes failure modes beyond simple collisions, and introduces the Non-Collision Failure (NCF) rate. Specifically, NCF aggregates operational breakdowns, including timeouts from excessive hovering, freezing in visually ambiguous areas without collision volumes, and severe wrong-way deviations. A high NCF indicates a critical misalignment between collision-avoidance heuristics and the agent's navigational intent.

\textbf{SFT \textit{w/} External APF Intervention.} Applying an external APF yields promising robustness metrics in familiar settings (e.g., Seen and Unseen Objects). However, beyond demanding auxiliary depth cameras and incurring extra computational overhead, this disjointed modular design introduces a severe semantic gap between the vision-language policy and the rigid physical intervention. This conflict inherently overrides visual navigation intentions, causing a significant drop in SR alongside an elevated NCF in Unseen Map scenarios. In contrast, AeroDPO internalizes robust collision-avoidance knowledge directly into an end-to-end model, ensuring superior generalization and computational efficiency without relying on external dependencies or heuristics.

\textbf{Ours \textit{w/o} Action Decoupling.} This variant demonstrates the risk of penalizing all control axes simultaneously, which induces a strictly over-conservative policy. While achieving exceptionally low CR, it catastrophically compromises efficiency. In the Unseen Map evaluation, its SPL regresses to 30.57\% with a peak NCF of 23.07\%, confirming the agent resorts to excessive hovering to evade penalties. AeroDPO effectively navigates this trade-off via decoupled interventions, maximizing both SR (49.16\%) and SPL (39.06\%) in the challenging UM split.

\textbf{Ours \textit{w/o} Data Curation.} Removing the VLM data curation leads to consistent declines in both SR and SPL, especially in generalization scenarios, as uncurated conflicting data severely confounds the spatial judgment of the model. This validates that filtering contradictory preference pairs remains essential for stable policy alignment.

\textbf{AeroDPO (Jetson INT8).} To validate applicability, we deploy an INT8 quantized AeroDPO model via \texttt{llama.cpp} on a Jetson Orin NX edge platform using hardware-in-the-loop simulation. This variant maintains a stable 770ms inference latency ($\sim$1.3 Hz). Table~\ref{tab:ablation} shows it surprisingly outperforms full-precision on Seen and Unseen Object splits, as the reduced precision of quantization inherently filters out spurious high-frequency visual noise \cite{lin2019defensive}, acting as an implicit regularizer that smooths state representations and mitigates erratic control jitter. Conversely, while its compressed capacity induces a performance drop on OOD Unseen Maps (44.99\% SR), it remarkably surpasses the full-precision baseline and all ablations. This highlights the profound robustness of the AeroDPO pipeline, demonstrating our automated preference alignment imparts resilient navigational priors, ensuring heavily quantized models maintain strong OOD generalization.

\section{Conclusion}

In this paper, we present AeroDPO, an ultra-lightweight yet robust end-to-end VLA paradigm that fundamentally challenges the parameter-heavy reliance of autonomous UAV navigation. We pioneer the empirical insight that high-fidelity perception fundamentally outweighs language reasoning scale, demonstrating that an agile 2B-parameter model can match or exceed massive 7B baselines while operating within an edge-deployable latency frontier. To rectify the severe robustness degradation inherent to pure BC in unmapped environments, AeroDPO introduces a zero-cost, automated offline preference alignment flywheel. By leveraging automated simulation state-rollback to isolate upstream causal errors, applying decoupled privileged interventions across spatial control axes, and filtering out visual ambiguities via an advanced vision-language curation loop, our agent successfully internalizes structural geometric boundaries without external online heuristics. Extensive closed-loop experiments validate that AeroDPO comprehensively outclasses state-of-the-art baselines, boosting the SR to 49.16\% on unmapped scenarios while drastically suppressing collision rates. Ultimately, this work establishes that embedding endogenous collision awareness via automated preference optimization delivers an elegant, robust, and highly scalable pathway toward fully autonomous aerial intelligence.

\bibliography{aaai2027}


\end{document}

%% file: table_perception_reasoning.tex
\begin{table*}[t]
\centering
\small
\setlength{\tabcolsep}{1.8pt}
\begin{tabular}{lcccccccccccccccccc}
\toprule
\multirow{2}{*}{\textbf{Model}} & \multirow{2}{*}{\textbf{Res.}} & \multirow{2}{*}{\textbf{Lat.}} & \multirow{2}{*}{\textbf{Mem.}} & \multicolumn{5}{c}{\textbf{Seen Map}} & \multicolumn{5}{c}{\textbf{Unseen Object}} & \multicolumn{5}{c}{\textbf{Unseen Map}} \\
\cmidrule(lr){5-9} \cmidrule(lr){10-14} \cmidrule(lr){15-19}
 & & & & NE$\downarrow$ & SR$\uparrow$ & OSR$\uparrow$ & SPL$\uparrow$ & CR$\downarrow$ & NE$\downarrow$ & SR$\uparrow$ & OSR$\uparrow$ & SPL$\uparrow$ & CR$\downarrow$ & NE$\downarrow$ & SR$\uparrow$ & OSR$\uparrow$ & SPL$\uparrow$ & \textbf{CR$\downarrow$} \\
\midrule
OpenVLA-7B & $224{\times}224$ & 0.45s & 17GB & \textbf{65.88} & 47.96 & \textbf{57.69} & 38.54 & - & \textbf{61.45} & 56.60 & \underline{64.86} & 46.61 & - & \textbf{67.42} & 37.58 & \textbf{52.92} & 28.22 & - \\
\midrule
Qwen3-vl-4B & $224{\times}224$ & 0.28s & 10GB & 74.75 & 45.20 & 53.17 & 36.07 & 37.59 & 74.67 & 54.37 & 58.51 & 46.15 & 27.50 & 86.12 & 33.82 & 43.11 & 27.41 & 56.89 \\
Qwen3-vl-4B & $224{\times}448$ & 0.30s & 10GB & \underline{66.96} & \textbf{49.15} & \underline{56.77} & \textbf{39.79} & \textbf{32.65} & \underline{62.49} & \textbf{60.89} & \textbf{67.73} & \textbf{51.40} & \textbf{21.14} & 77.28 & \underline{40.08} & 48.64 & \underline{32.87} & \underline{54.80} \\
Qwen3-vl-2B & $224{\times}448$ & 0.20s & 5GB & 69.82 & 47.53 & 54.94 & 37.15 & 35.40 & 67.36 & 58.35 & 64.55 & 48.78 & \underline{25.12} & 88.33 & 35.18 & 41.86 & 28.28 & 56.47 \\
\midrule
Qwen3-vl-2B & $384{\times}768$ & 0.24s & 5GB & 69.14 & \underline{48.59} & 55.57 & \underline{38.91} & \underline{34.27} & 64.70 & \underline{59.78} & 63.59 & \underline{50.08} & 25.59 & \underline{73.90} & \textbf{41.23} & \underline{50.42} & \textbf{34.46} & \textbf{46.76} \\
\bottomrule
\end{tabular}%
\caption{Performance comparison of varying parameter scales and input resolutions on the TravelUAV benchmark. This table evaluates the baseline OpenVLA-7B against lightweight Qwen3-VL variants across familiar and out-of-distribution environments. The metrics capture navigation task completion, computational efficiency, and flight safety as indicated by the Collision Rate (CR). Bold and underlined values represent the best and second-best results respectively.}
\label{tab:perception_reasoning}
\end{table*}

%% file: table_seen.tex
\begin{table*}[tbp]
\centering
\small
\setlength{\tabcolsep}{4pt}
\begin{tabular}{llcccccccccccc}
\toprule
\multirow{2}{*}{Method} & \multirow{2}{*}{Guide} & \multicolumn{4}{c}{\textbf{Full}} & \multicolumn{4}{c}{\textbf{Easy}} & \multicolumn{4}{c}{\textbf{Hard}} \\
\cmidrule(lr){3-6} \cmidrule(lr){7-10} \cmidrule(lr){11-14}
 & & NE$\downarrow$ & SR$\uparrow$ & OSR$\uparrow$ & SPL$\uparrow$ & NE$\downarrow$ & SR$\uparrow$ & OSR$\uparrow$ & SPL$\uparrow$ & NE$\downarrow$ & SR$\uparrow$ & OSR$\uparrow$ & SPL$\uparrow$ \\
\midrule
TravelUAV \cite{wang2025towards} & L1 & 98.66  & 17.45 & 48.87 & 15.76 & 66.40  & 20.26 & 51.23 & 18.10 & 138.04 & 14.02 & 45.98 & 12.90 \\
NavFoM \cite{zhang2026embodied}  & L1 & 93.05  & 29.17 & 49.24 & 25.03 & 58.98  & 32.91 & 53.16 & 27.87 & 143.83 & 23.58 & 43.40 & 20.80 \\
LongFly \cite{jiang2025longfly} & L1 & \underline{60.02} & 36.39 & \underline{65.87} & 31.07 & \underline{38.10} & 38.52 & \textbf{71.90} & 31.24 & \underline{85.20} & 33.94 & \underline{58.94} & 30.88 \\
NeuroKalman \cite{tang2026mitigating} & L1 & 71.56 & 25.86 & 58.73 & 22.43 & 42.70 & 30.52 & 62.70 & 25.86 & 105.07 & 20.11 & 53.90 & 18.21 \\
AeroVLA \cite{xu2026aerovla} & FD & 65.88 & 47.96 & 57.69 & 38.54 & 43.76 & 49.30 & 61.30 & 37.14 & 93.16 & 46.30 & 53.23 & \underline{40.26} \\
\midrule
Baseline - SFT Model & FD & 69.14 & \underline{48.59} & 55.57 & \underline{38.91} & 45.16 & \underline{50.32} & 59.00 & \underline{37.84} & 98.71 & \underline{46.46} & 51.34 & 40.22 \\
\textbf{Ours} & FD & \textbf{53.30} & \textbf{60.93} & \textbf{67.49} & \textbf{48.47} & \textbf{36.03} & \textbf{63.73} & \underline{70.75} & \textbf{47.13} & \textbf{74.60} & \textbf{57.48} & \textbf{63.47} & \textbf{50.11} \\
\bottomrule
\end{tabular}
\caption{Comparison on the Test Seen Set. SR, OSR, and SPL are reported in percentage (\%). Bold and underline denote the best and second-best results. L1 and FD denote L1 Assist and Fuzzy Direction, respectively.}
\label{tab:comparison_seen}
\end{table*}

%% file: table_unseen.tex
\begin{table*}[t]
\centering
\small
\setlength{\tabcolsep}{3pt}
\begin{tabular}{cllcccccccccccc}
\toprule
\multirow{2}{*}{\textbf{Split}} & \multirow{2}{*}{Method} & \multirow{2}{*}{Guide} & \multicolumn{4}{c}{\textbf{Full}} & \multicolumn{4}{c}{\textbf{Easy}} & \multicolumn{4}{c}{\textbf{Hard}} \\
\cmidrule(lr){4-7} \cmidrule(lr){8-11} \cmidrule(lr){12-15}
 & & & NE$\downarrow$ & SR$\uparrow$ & OSR$\uparrow$ & SPL$\uparrow$ & NE$\downarrow$ & SR$\uparrow$ & OSR$\uparrow$ & SPL$\uparrow$ & NE$\downarrow$ & SR$\uparrow$ & OSR$\uparrow$ & SPL$\uparrow$ \\
\midrule
\multirow{9}{*}{\textbf{UO}}
 & TravelUAV \cite{wang2025towards} & L1 & 118.11 & 22.42 & 46.90 & 20.51 & 86.12  & 24.40 & 49.28 & 22.03 & 134.03 & 21.43 & 45.71 & 19.75 \\
 & NavFoM \cite{zhang2026embodied}  & L1 & 108.04 & 29.83 & 47.99 & 27.20 & 70.51  & 32.54 & 50.72 & 29.54 & 133.01 & 28.03 & 46.18 & 25.64 \\
 & LongFly \cite{jiang2025longfly} & L1 & 66.74 & 43.87 & 64.56 & 38.39 & 54.84 & 38.01 & 56.84 & 31.36 & \underline{57.07} & 50.25 & \textbf{74.16} & 45.27 \\
 & NeuroKalman \cite{tang2026mitigating} & L1 & 71.01 & 32.48  & 60.82 & 28.50  & \underline{44.50} & 42.50  & \underline{66.50} & 37.37  & 84.50 & 27.50  & 58.00  & 24.50  \\
 & AeroVLA \cite{xu2026aerovla} & FD & \underline{61.45} & 56.60 & \underline{64.86} & 46.61 & 45.72 & 56.94 & 64.11 & 43.76 & 69.27 & 56.43 & 65.24 & 48.03 \\
 \cmidrule(lr){2-15}
 & Baseline - SFT Model & FD & 64.70 & \underline{59.78} & 63.59 & \underline{50.08} & 44.82 & \underline{59.33} & 65.07 & \underline{47.53} & 74.59 & \underline{60.00} & 62.86 & \underline{51.35} \\
 & \textbf{Ours} & FD & \textbf{47.31} & \textbf{67.41} & \textbf{71.55} & \textbf{57.54} & \textbf{33.31} & \textbf{67.47} & \textbf{72.73} & \textbf{54.27} & \textbf{54.27} & \textbf{67.38} & \underline{70.95} & \textbf{59.16} \\
\midrule
\multirow{9}{*}{\textbf{UM}}
 & TravelUAV \cite{wang2025towards} & L1 & 138.80 & 4.18  & 20.77 & 3.84  & 102.94 & 4.63  & 22.82 & 4.24  & 189.46 & 3.53  & 17.88 & 3.28  \\
 & NavFoM \cite{zhang2026embodied}  & L1 & 125.10 & 6.30  & 18.95 & 5.68  & 102.41 & 6.77  & 20.07 & 6.04  & 170.58 & 5.36  & 15.71 & 4.97  \\
 & LongFly \cite{jiang2025longfly} & L1 & 108.32 & 11.27 & 30.27 & 9.32 & 78.56 & 12.96 & 34.31 & 10.32 & 148.10 & 9.02 & 24.88 & 7.98 \\
 & NeuroKalman \cite{tang2026mitigating} & L1 & 100.32 & 8.34  & 34.15 & 7.12  & 69.50 & 9.15  & 38.50 & 7.50  & 140.00 & 7.20  & 28.00  & 6.50  \\
 & AeroVLA \cite{xu2026aerovla} & FD & \underline{67.42} & 37.58 & \underline{52.92} & 28.22 & \textbf{44.99} & 41.89 & \underline{58.47} & 29.72 & \underline{99.11} & 31.49 & \underline{45.09} & 26.11 \\
 \cmidrule(lr){2-15}
 & Baseline - SFT Model & FD & 73.90 & \underline{41.23} & 50.42 & \underline{34.46} & 51.96 & \underline{45.81} & 55.97 & \underline{37.13} & 104.90 & \underline{34.76} & 42.57 & \underline{30.69} \\
 & \textbf{Ours} & FD & \textbf{66.96} & \textbf{49.16} & \textbf{57.72} & \textbf{39.06} & \underline{48.70} & \textbf{52.23} & \textbf{61.32} & \textbf{39.48} & \textbf{92.76} & \textbf{44.84} & \textbf{52.64} & \textbf{38.46} \\
\bottomrule
\end{tabular}
\caption{Zero-Shot Generalization on OOD Environments. UO and UM denote Unseen Object and Unseen Map. SR, OSR, and SPL are reported in percentage (\%). Bold denotes the best results, and underline indicates the second best. In the Guide column, L1 denotes L1 Assist, and FD denotes Fuzzy Direction.}
\label{tab:comparison_ood}
\end{table*}

%% file: table_ablation.tex
\begin{table*}[t]
\centering
\small
\setlength{\tabcolsep}{4.5pt}
\begin{tabular}{l cccc cccc cccc}
\toprule
\multirow{2}{*}{Model Variant} & \multicolumn{4}{c}{\textbf{Seen}} & \multicolumn{4}{c}{\textbf{Unseen Object}} & \multicolumn{4}{c}{\textbf{Unseen Map}} \\
\cmidrule(lr){2-5} \cmidrule(lr){6-9} \cmidrule(lr){10-13}
 & SR$\uparrow$ & SPL$\uparrow$ & CR$\downarrow$ & NCF$\downarrow$ 
 & SR$\uparrow$ & SPL$\uparrow$ & CR$\downarrow$ & NCF$\downarrow$ 
 & SR$\uparrow$ & SPL$\uparrow$ & CR$\downarrow$ & NCF$\downarrow$ \\
\midrule
SFT Baseline                              & 48.59 & 38.91 & 34.27 & 16.15 & 59.78 & 50.08 & 25.59 & \textbf{14.00} & 41.23 & 34.46 & 46.76 & \textbf{7.72} \\
SFT \textit{w/} External APF Intervention & 59.52 & 46.69 & \underline{20.24} & 18.90 & \textbf{70.59} & \textbf{60.72} & \textbf{11.76} & 16.06 & 43.42 & 34.11 & 36.74 & 14.09 \\
Ours \textit{w/o} Action Decoupling       & \underline{61.57} & 46.41 & \textbf{19.25} & 17.55 & 67.09 & 53.31 & \underline{12.56} & 18.60 & 43.74 & 30.57 & \textbf{28.50} & 23.07 \\
Ours \textit{w/o} Data Curation           & 58.11 & 47.03 & 29.55 & \textbf{11.35} & 62.32 & 51.53 & 20.99 & 15.11 & \underline{44.99} & \underline{36.09} & 38.94 & 10.96 \\
\midrule
\textbf{AeroDPO (Ours)}                   & 60.93 & \underline{48.47} & 26.80 & 11.77 & 67.41 & 57.54 & 16.69 & \underline{14.47} & \textbf{49.16} & \textbf{39.06} & \underline{35.91} & \underline{10.33} \\
AeroDPO (Jetson INT8)                     & \textbf{62.62} & \textbf{50.57} & 25.32 & \underline{11.57} & \underline{70.11} & \underline{59.59} & 13.99 & 15.74 & \underline{44.99} & 35.59 & 41.86 & 12.22 \\
\bottomrule
\end{tabular}
\caption{Comprehensive ablation studies across Seen, Unseen Object, and Unseen Map splits. We evaluate Success Rate (SR), Success weighted by Path Length (SPL), Collision Rate (CR), and the Non-Collision Failure (NCF) rate, which aggregates Time Out, Stuck, and Wrong Way errors. Bold and underlined values represent the best and second-best results respectively.}
\label{tab:ablation}
\end{table*}